\documentclass[preprint,review,12pt]{elsarticle}

\usepackage{amssymb}
\usepackage[numbers]{natbib}
\usepackage{graphicx} 
\usepackage{algorithm}
\usepackage{algpseudocode}

\usepackage{times} 
\usepackage{amsmath} 

\journal{Engineering Applications of Artificial Intelligence}

\begin{document}
	
	\begin{frontmatter}
		
		
		
		\title{Catching Spinning Table Tennis Balls in Simulation with End-to-End Curriculum Reinforcement Learning}
		
		
		\author[1]{Xiaoyi Hu}
		
		\author[1]{Yue Mao}
		
		\author[1]{Gang Wang}

		\author[1]{Qingdu Li}
		\author[2]{Jianwei Zhang}
		\author[1]{Yunfeng Ji \corref{cor1}}
		\ead{ji_yunfeng@usst.edu.cn} 
		\cortext[cor1]{Corresponding author}
		
		\address[1]{University of Shanghai for Science and Technology, Shanghai 200093,China}
		\address[2]{University of Hamburg, Hamburg 20148, Germany}
		
		\tnotetext[funding]{
			This research was funded by the Natural Science Foundation of China under Grant 62403319.
		}
		
		\begin{abstract}
			The game of table tennis is renowned for its extremely high spin rate, but most table tennis robots today struggle to handle balls with such rapid spin. To address this issue, we have contributed a series of methods, including: 1. Curriculum Reinforcement Learning (RL): This method helps the table tennis robot learn to play table tennis progressively from easy to difficult tasks.
			2. Analysis of Spinning Table Tennis Ball Collisions: We have conducted a physics-based analysis to generate more realistic trajectories of spinning table tennis balls after collision.
			3. Definition of Trajectory States: The definition of trajectory states aids in setting up the reward function.
			4. Selection of Valid Rally Trajectories: We have introduced a valid rally trajectory selection scheme to ensure that the robot's training is not influenced by abnormal trajectories.
			5. Reality-to-Simulation (Real2Sim) Transfer: This scheme is employed to validate the trained robot’s ability to handle spinning balls in real-world scenarios. With Real2Sim, the deployment costs for robotic reinforcement learning can be further reduced. Moreover, the trajectory-state-based reward function is not limited to table tennis robots; it can be generalized to a wide range of cyclical tasks. To validate our robot’s ability to handle spinning balls, the Real2Sim experiments were conducted. For the specific video link of the experiment, please refer to: https://www.youtube.com/watch?v=uAx7eYAGylc.
			
		\end{abstract}

		
		
		
	\end{frontmatter}
	
	
	\section{INTRODUCTION}
	
	In all ball sports, table tennis (Ping-Pong) is renowned for its spin. The high spin not only changes the trajectory of the table tennis ball but also makes it more difficult to hit it back to the opponent's table. Due to the high cost involved in measuring the spin speed of table tennis balls, predicting the trajectory of spinning table tennis balls and formulating corresponding hitting policies have become extremely challenging tasks. Therefore, most table tennis robots today are unable to cope with high-speed spinning table tennis balls.
	
	Despite the difficulty in returning spinning table tennis balls, there has been a significant amount of research on the hitting policies for table tennis robots. At present, there are two methods to implement the hitting policy: the model-based hitting policy and the data-driven hitting policy. The main feature of the model-based hitting policy is to predict the table tennis trajectory by using the physical model and state estimation method. Then, the hitting action is made according to the prediction \cite{kocc2018online}, \cite{liu2013racket}, \cite{zhang2010visual}. Although this policy has excellent interpretability, unprecise physical properties may degrade its performance. The data-driven policy utilize playing data to improve the performance of the table tennis robot. This training method encompasses supervised learning to improve the accuracy of trajectory prediction \cite{wu2020futurepong}, \cite{tebbe2020spin}, \cite{gomez2020real} and optimizing robot performance with control theory \cite{zhou2020design} . These methods mentioned above often require significant costs for data collection and labeling. Moreover, policies based on labeled data may not optimal. Therefore, a better solution is to enable robots to actively collect data and use it for self-updates and improvements.

	In recent years, with the rapid development of reinforcement learning (RL), using reinforcement learning in robotics has gradually become feasible. However, since this learning method requires a significant amount of trial and error to collect data, ensuring safety has been a concern \cite{yuan2022safe}, \cite{brunke2022safe}.	
	To ensure the safety of table tennis robots in RL, many researchers have proposed various solutions. \cite{buchler2022learning} proposed a robot driven by pneumatic muscles so that the robot can perform trials and errors without causing any damage even colliding  with itself or external objects. To ensure that the actions performed by the trained robots are safe, \cite{tebbe2021sample} proposed that RL should only be used to determine the posture and velocity of the racket, while the control elements that may cause unexpected accidents, such as the racket position and joint angle, should be implemented by other safer methods.

	Despite addressing safety concerns, the extremely low data reuse in reinforcement learning leads to high training costs for robots in real-world scenarios \cite{kalashnikov2018scalable}, \cite{akkaya2019solving}. This not only wastes lots of time but also causes fatigue in the mechanical structure of the robot\cite{hofer2021sim2real}. Compared to training robots in reality, training robots in a simulation can significantly expedite the sampling efficiency through multi-environment. Additionally, with the help of Sim2Real, it becomes possible to transfer the learned behaviors of robots from the simulation to the real-world environment \cite{amini2020learning}. This approach accelerates data collection and reduces costs, making it a promising solution for safe and efficient robot learning in real world. 
	
	Considering the factors mentioned above, the main focus of this paper is on how to train a table tennis robot in a simulation environment that can effectively catch spinning balls. Additionally, the feasibility of this approach will be demonstrated through Real2Sim experiments, which involve transferring the learned behaviors from simulation to the real world. The goal is to show that the training in the simulated environment can effectively equip the robot to handle spinning balls and perform well in real-world settings.
	
	1. Curriculum Reinforcement Learning: Curriculum reinforcement learning is an special method in reinforcement learning that involves dividing a training task into multiple subtasks\cite{wang2023curriculum}, \cite{narvekar2020curriculum}. By employing this method, the inherent difficulty of the training task is reduced, leading to enhanced performance of the learning agent. Therefore, we introduce a new end-to-end table tennis robot training method on the basis of curriculum reinforcement learning. The curriculum reinforcement learning method employed in this study requires the robot to progressively learn three tasks: catching the ball, hitting the ball, and hitting the ball to the target point.
	
	2. Analysis of Spinning Table Tennis Ball Collisions: When using Sim2Real for robot reinforcement learning, the performance of the robot is often affected by the accuracy of the simulation environment. In many simulation environments, trajectory deviations caused by high-speed spinning table tennis ball collisions are not considered. To address this issue, this study analyzes the velocity changes before and after the collision of spinning table tennis balls based on a physics model. The physics model is integrated with the existing simulation environment to enhance the accuracy of simulating spinning ball collisions.
	
	3. Definition of Trajectory States: Since training an end-to-end table tennis robot solely using curriculum learning cannot fully achieve the target, a trajectory state definition method for table tennis rally is introduced in this study. During training for different tasks, the reward function varies based on the current trajectory state of the table tennis robot. Therefore, we propose a matrix-based approach to define the reward function, making the reward function for curriculum reinforcement learning more intuitive to comprehend.
	
	4. Selection of Valid Rally Trajectories: To improve the learning efficiency and reduce the impact of irrelevant samples on the agent, this paper constructs a scheme to automatically generate valid spinning table tennis ball trajectories according to the definition of trajectory states.
	
	5. Reality-to-Simulation (Real2Sim) Transfer: To verify the feasibility of the proposed method, real-world table tennis ball trajectories are transmitted into the simulationThen, the table tennis robot will be required to catch the spinning table tennis ball in the simulation environment to demonstrate its ability to handle spinning balls.

	These contributions collectively enhance the training of the table tennis robot, enabling it to successfully respond to opponents' attacks and accurately hit the table tennis ball to the target point. The Real2Sim experiments have successfully demonstrated the feasibility of our scheme.
	
	In the aforementioned innovations, we believe these contributions are not limited to table tennis robots. Other research in robotic reinforcement learning can also benefit from our findings. To address the current challenge of high deployment costs in reinforcement learning, the Real2Sim technique proposed in this paper can be employed. This approach allows for the pre-assessment of a real robot's decision-making based on actual data, enabling further adjustments to the agent's strategy.

	Additionally, the definition of trajectory states and the corresponding reward functions can be applied to a wider range of periodic tasks. In this study, we treat the table tennis rally as a cycle, where the trajectory states represent key points within that cycle. By designing distinct reward functions for these key points, we can enhance the robot's performance. For example, walking is a typical cyclical task. When the walking gait is viewed as a cycle, the standing phase and swinging phase of the legs can be regarded as key points. Here. the "key points" refer to the trajectory states within the table tennis rally.  Therefore, if the concept of trajectory states is generalized, it could provide a novel solution for addressing various cyclical tasks.

	\section{CONFIGURATION OF OUR ROBOT SYSTEM}
	
\begin{figure}
		\centering
		\includegraphics[scale=0.5]{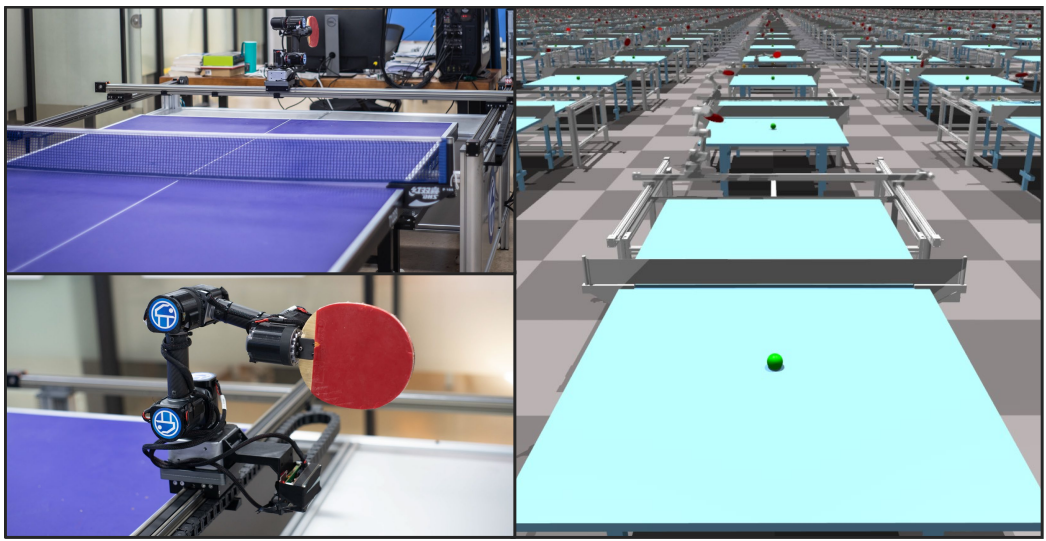}
		\caption{Comparison between the simulation (Isaac Gym) and the actual mechanical structure of the table tennis robot.}
		\label{fig:robotsystem}
	\end{figure}

	The mechanical structure of the simulated robot in this paper is consistent with the table tennis robot developed by our team in \cite{ji2021model}. The comparisons between the real-world robot and the simulated robot are shown in Fig. \ref{fig:robotsystem}. Due to the disparity between the simulation and the reality, we use several techniques to close the gap. For example, noise and latency are introduced to the perception system of the simulated robot to replicate the interference that a real robot may experience. In addition, randomization is used to improve the ability of the model to generalize for difficult-to-measure physical properties. However, the complexity of the simulation environment may lead to an increase in the training difficulty. Thus, Isaac Gym \cite{makoviychuk2021isaac} is selected as the simulation environment during the training. By operating thousandss of environments in parallel on a single GPU, Isaac Gym significantly enhances the speed of sampling data for RL, making the proposed scheme feasible.
	
	To simplify the task, we ignore the serving phase of table tennis and mainly focus on the human-robot rally. However, table tennis is a sparse reward task, and directly training the robot according to the scoring mechanism may prevent the robot from learning the task objective because of a lack of positive and negative feedback. Therefore, in this paper, we divide the human-robot rally process into multiple trajectory states. Moreover, the  reward functions may change according to the current trajectory state of table tennis to reduce the learning difficulty. 	To facilitate the collection of training data, all rally trajectories have been categorized into two groups: valid rally trajectories and invalid rally trajectories. For simplicity, we define a valid rally trajectory as one that can cross the net and its first contact point is on the court of the robot and an abnormal rally trajectory as one that does not. This categorization is due to the rules of table tennis, which specify that during a rally, the table tennis ball must first make contact with the opponent's table before the opponent can strike it. Most random rally trajectories are invalid rally; thus, the robot cannot learn anything from them, which decreases the sampling efficiency. Therefore, in this paper, a method for obtaining valid rally trajectories efficiently in parallel environments is developed. In the simulation environment, external factors such as wind are not considered, ensuring that a table tennis ball with identical initial velocity, spin, and position will generate the same trajectory. Furthermore, to account for the impact of air drag and Magnus force on the ball's trajectory, we record the coefficients associated with these two forces. It is worth noting that the elastic coefficient of the table tennis table, along with the coefficients of sliding and rolling friction, exclusively influences the trajectory post-collisions and does not interfere with the valid rally trajectory. As a result, by capturing and utilizing the mentioned data, we can faithfully reproduce the valid rally trajectory.

	\subsection{The Definition of Trajectory State} 
	
\begin{figure}[tbph]
		\centering
		\includegraphics[scale=0.7]{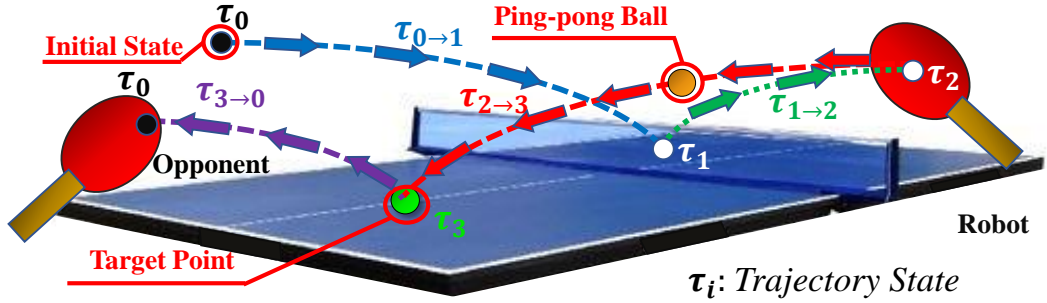}\vspace{0cm}
		\caption{Trajectory state loop for table tennis.}\vspace{0cm}
		\label{fig:Trajectory state loop}
	\end{figure}

	The trajectory state
	$
	\boldsymbol{\tau} = ~\{~\tau_0,~\tau_{0\rightarrow1},~\tau_1,~\tau_{1\rightarrow2},~\tau_2, ~\tau_{2\rightarrow3},~ \tau_3,~ \tau_{3\rightarrow0}\}
	$ can be separated into two categories: the instantaneous trajectory state $\tau_i$ for sparse rewards in the reward function and the continuous trajectory state $\tau_{i\rightarrow j}$ for dense rewards in the reward function. All the abovementioned trajectory states can only transition from one state to the next state step by step. The instantaneous trajectory state $\tau_i$ and the continuous trajectory state $\tau_{i\rightarrow j}$ of table tennis may form a complete rally. The definitions of each trajectory state are shown in TABLE \ref{table_trajectory_state}.
	\begin{table}
		\vspace{0.3cm}
		\caption{the trajectory state of table tennis}
		\label{table_trajectory_state}
		\begin{center}
			\vspace{-0.5cm}
			\begin{tabular}{c|c}
				\hline
				Symbol & Name                                                                                                                                                 \\ \hline \hline
				$\tau_0$	& Initial state of the ball                   \\ \hline
				$\tau_{0\rightarrow1}$	& Opponent hitting period                     \\ \hline
				$\tau_1$	& The ball collides with the robot's court    \\ \hline
				$\tau_{1\rightarrow2}$	& Robot catching period                       \\ \hline
				$\tau_2$	& The ball collides with the robot's racket   \\ \hline
				$\tau_{2\rightarrow3}$	& Robot hitting period                        \\ \hline
				$\tau_3$	& The ball collides with the opponent's court \\ \hline
				$\tau_{3\rightarrow0}$	& Opponent catching period  \\\hline                  
			\end{tabular}
		\end{center} 
		\vspace{-0.5cm}\end{table}
	According to the above trajectory state definition, the human-robot rally can be described as $\tau_0\rightarrow\tau_{0\rightarrow1}\rightarrow\tau_1\rightarrow\tau_{1\rightarrow2}\rightarrow\tau_2\rightarrow\tau_{2\rightarrow3}\rightarrow\tau_3\rightarrow\tau_{3\rightarrow0}\rightarrow\tau_0\rightarrow... $, which can continue to exist until one side fails to catch the ball or fails to hit the ball to the opponent's court. The schematic of this loop is shown in Fig. \ref{fig:Trajectory state loop}.
	
	\subsection{Collision Model of Spinning Table Tennis Ball} 
	The table tennis collision model developed in this paper focuses on describing the post-collision trajectory of the spinning table tennis ball. In contrast to the original table tennis ball trajectories in the simulation, this model additionally considers the state changes caused by the strong spin of the table tennis ball before and after collisions. In a table tennis rally, its trajectory can suddenly change twice: when the table tennis ball collides with the table (trajectory state  $\tau_{1}$) and when it collides with the paddle (trajectory state 	 $\tau_{2}$).
	
\begin{figure}[tbph]
		\centering
		\includegraphics[scale=2]{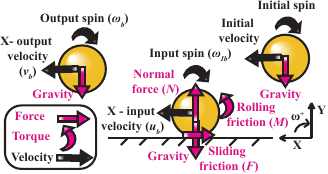}\vspace{0cm}
		\caption{Analysis of the table tennis ball's state before and after collision.}\vspace{0cm}
		\label{fig:spin}
		
	\end{figure}

	When the table tennis ball contacts the paddle, the paddle imparts an impulse, altering the ball's velocity and spin. This not only changes the ball's movement trajectory but also affects the robot's ability to hit the ball onto the opponent's table. Similarly, the collision between the table tennis ball and the table follows a similar pattern, except that no additional impulse is applied. This paper primarily analyzes the velocity changes with physical model when the table tennis ball collides with the table and the racket. This approach not only enhances the simulation's accuracy but also enables the robot to better adapt to the characteristics of the table tennis ball with strong spin. The analysis of a highly spinning table tennis ball's state before and after collision is shown in Fig.\ref{fig:spin}.
	
	Based on the conservation of momentum and kinetic energy, the following equations exist:
	\begin{equation}
		\left\{ \begin{array}{lr}
			\frac{1}{2}(m_{b}v_{b}^2+I_{b}\omega_{b}^2)=\frac{1}{2}(m_{b}u_{b}^2+I_{b}\omega_{Ib}^2)-\int{F(s)ds}-\int{M(\theta)d\theta}\\
			m_bu_b=m_bv_b+\int{f(t)}dt\\
			I_b\omega_b=I_b\omega_b+\int{}M(t)dt\end{array}
		\right .	
	\end{equation}
	Here $m_b$ and $I_b$ represent the mass and rotational inertia of the ball, while $u_b$, $\omega_{Ib}$, $v_b$, and $\omega_b$ denote the linear and angular velocity before collision, linear and angular velocity after collision. After simplification, the equations are as follows:
	\begin{equation}
		\left\{ \begin{array}{lr}
			\omega_{b} = \omega_{Ib} + \frac{\int{}M(t)dt}{I_b}\\
			v_b = \frac{2\int{}f(s)ds+2\int{}M(\theta)d\theta-(\omega_b+\omega_{Ib})\int{}M(t)dt}{\int{}f(t)dt}-u_b\\
		\end{array}
		\right .
	\end{equation}
	
	In the above equations, most values can be directly obtained from the simulation environment, while $\int{}f(s)ds$, $\int{}M(\theta)d\theta$, $\int{}M(t)dt$, and $\int{}f(t)dt$ can be replaced by sums of discrete data. From the above equations, it is evident that when the ball possesses a significant counterclockwise angular velocity, the linear speed will be enhanced after the collision due to the effect of friction torque. Conversely, when the ball has a strong clockwise angular velocity, the linear speed after the collision can even be completely opposite to the linear speed before the collision. Both of these scenarios are consistent with real-world situations. By using this collision model, the trajectory of the table tennis ball after collision can be obtained and applied to the training of the robot.

	\subsection{ Valid Rally Trajectory Generation in Parallel Environments}
	\begin{figure*}[tbph]
		\centering
		\includegraphics[scale=0.3]{Valid\_trajectory.pdf}\vspace{-0cm}\hspace{-0cm}
		\caption{Valid rally trajectory generation in parallel environments.}\vspace{-0cm}
		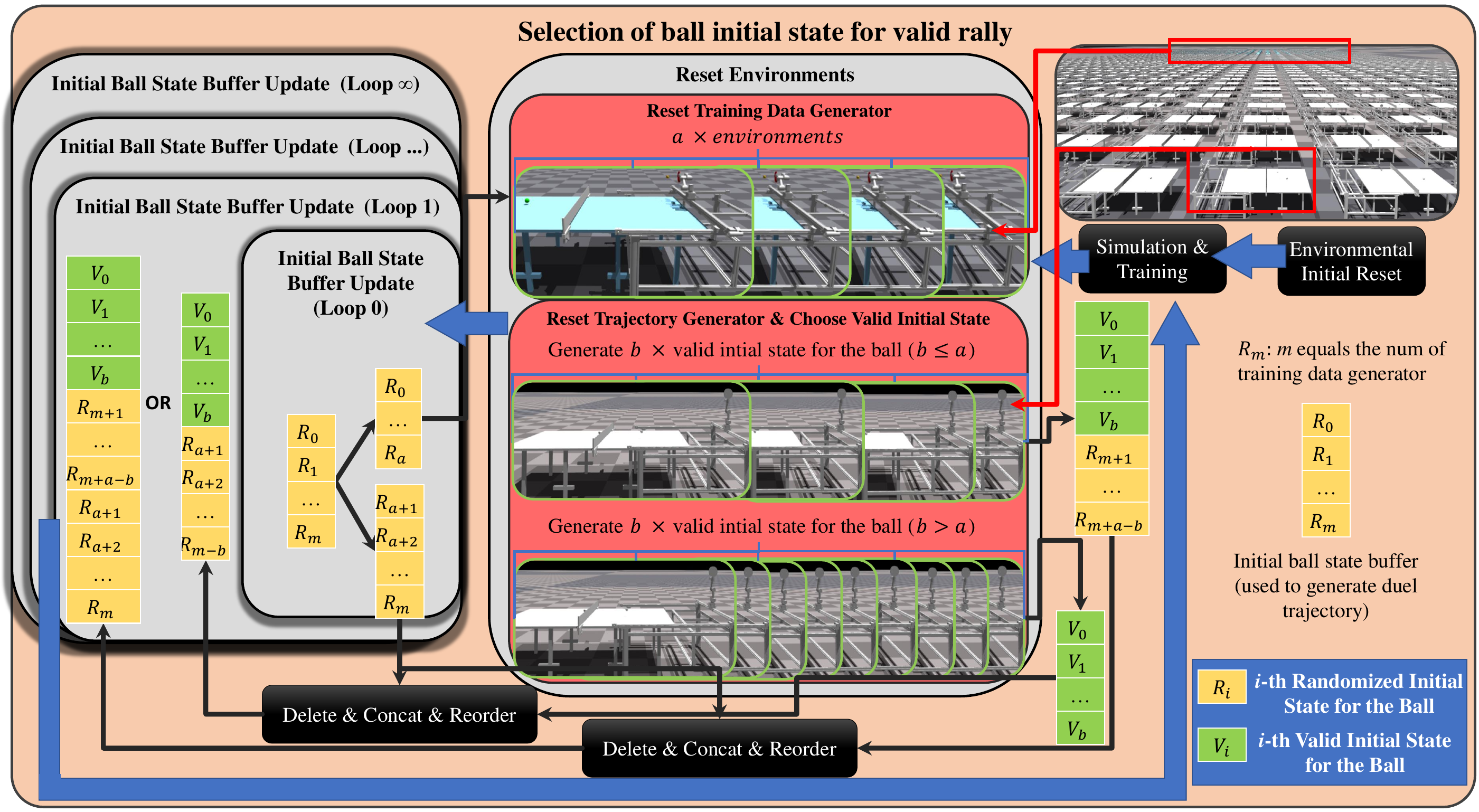
\label{fig:Valid_trajectory.pdf}
	\end{figure*}
	The trajectories that satisfy\ $\tau_0\rightarrow\tau_{0\rightarrow1}\rightarrow\tau_1$ are defined as valid rally trajectories, i.e., the ball starts from the initial position and only collides with the robot court. For the convenience of recording the rally trajectory, all the parallel environments are divided into two parts: the rally trajectory generator for producing valid rally trajectories and the training data generator for training the robot. In the rally trajectory generator, the initial position, velocity, and spin of the table tennis ball, as well as the coefficients for air drag and Magnus force, are randomly set to explore valid rally trajectories.In the training data generator, the initial data of the valid rally trajectory found by the trajectory generator are used so that the robot can be trained without interference from invalid data. To improve the sampling efficiency, the trajectory generator and the training data generator will run in parallel. Since the rally trajectory generator needs to conduct a large number of random trials to generate valid rally trajectories, the consumption speed of the training data generator will be faster than that of the trajectory generator. Therefore, a buffer with a length equal to the number of training data generators is established to store the initial data of the valid rally trajectory. 
	Every time the environment is reset, the initial data of the valid trajectory will be stored in the buffer, and when there are no available data in the buffer, random initial data are used for replacement. The details of this process are shown in Fig. \ref{fig:Valid_trajectory.pdf}.
	
	\begin{figure*}[tbph]
		\centering
		\vspace{0.25cm}
		\includegraphics[scale=0.7]{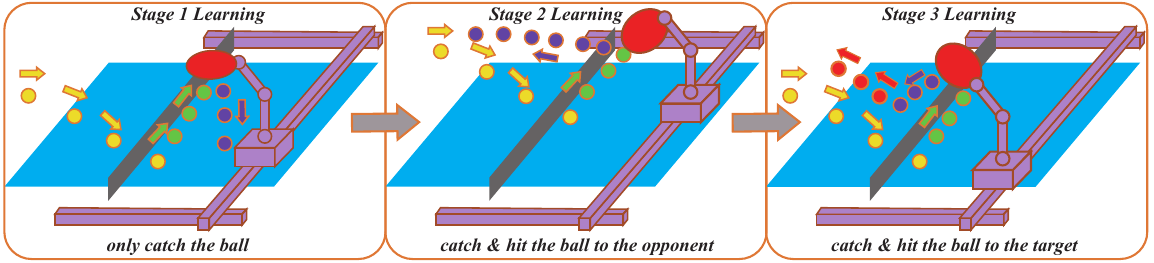}\vspace{0cm}
		\caption{Curriculum reinforcement learning for table tennis robot.}\vspace{0cm}
		\label{fig:curriculum}
		\vspace{-0.5cm}
	\end{figure*}
	
	\section{TRAINING HITTING POLICY IN SIMULATION}	
	During the training, agents begin with random behaviors and strive to achieve the greatest reward according to the reward function. However, the causal reward function may force agents to stray from the training objective, and the agent may engage in odd behavior to pursue the maximum reward. This may results that the policy with the highest reward is not the best policy that humans expect. Moreover, if the learning task is too difficult, the policy may become stuck and cease to learn. To address this issue, we employed curriculum reinforcement learning.
	To facilitate training, robot's learning process has been divided into three stages. Human prior knowledge is introduced by setting the reward function for each stage that the robot must complete in turn. The general process of the curriculum reinforcement learning is illustrated in Fig. \ref{fig:curriculum}.
	
	\subsection{Reward function $^sr$ in curriculum learning}
	
	When humans learn to play table tennis, we assume that they progress from easier to more difficult tasks. The hypothetical learning process we followed includes: hitting the ball, returning the ball, and finally aiming to land the ball on the opponent's side of the table. By structuring the subtasks in this manner, we aim to better implement curriculum learning to improve the robot's performance in table tennis.
	By considering the inspired by the assumption, we can design a reward function that guides the table tennis robot to gradually learn and improve its performance in a step-by-step manner.
	With curriculum learning, the anticipated skills of the robot are predefined, and the complexity of the task ranges from simple to challenging. This causes the policy to be closer to the ideal policy expected by humans and eliminates the demand for demonstration data. However, it is a significant challenge to allow the robot to perform well in each subtask. To simplify the problem, a sparse reward can be transformed into a dense reward based on the trajectory state $\tau$ in rally trajectories. 
	
	The reward function $^sr$ for this table tennis robot is:
	\begin{equation*}
		\begin{matrix}[^1r,^2r,^3r] =  \boldsymbol{\tau} R\\
			^sr = ^{i}r\end{matrix}		
	\end{equation*}
	\vspace{-0.8cm}
	\begin{flalign}
		&\ R=&	\nonumber
	\end{flalign}
	\vspace{-1cm}
	\begin{equation}
		\begin{matrix}
			&\begin{bmatrix} 0 & 0 &0 \\{a_{21}/{\left(1+d_{rb}^2\right)^2}} & a_{22}/{\left(1+d_{rb}^2\right)^2} & a_{23}/{\left(1+d_{rb}^2\right)^2} \\ {a_{31}} &a_{32} & a_{33} \\{a_{41}/{\left(1+d_{rb}^2\right)^2}} &a_{42}/{\left(1+d_{rb}^2\right)^2} &{a_{43}}/{\left(1+d_{rb}^2\right)^2}\\ a_{51} & a_{52}+v_{rhb_x} &a_{53} \\ 0& 0& a_{63}/{\left(1+d_{bt}^2\right)^2}\\ 0&0&a_{73}+b_{73}/{\left(1+e_{lt}^2\right)^2} \\0&0  &0
				\\	
			\end{bmatrix}\\
			&\boldsymbol{\tau} =\begin{bmatrix} 				\tau_0 &	\tau_{0\rightarrow1} &	\tau_1& 			\tau_{1\rightarrow2}	& 
				\tau_2	& 
				\tau_{2\rightarrow3}	& 
				\tau_3	& 
				\tau_{3\rightarrow0}	&    \end{bmatrix}\\
		\end{matrix}
	\end{equation}
	In the equation above, $R$ represents the reward coefficient matrix for each subtask in different stages, and $\boldsymbol{\tau}$ involves all the trajectory states in a valid rally. 
	$a_{jk}$ and $b_{jk}$ are the hyperparameters in the reward function, where $j$ represents the current trajectory state, and $k$ indicates the current training stage.
	When calculating the reward, $\boldsymbol{\tau}$ is a one-hot vector, representing the current trajectory state $\tau$ of the ball.  $^1r$, $^2r$, and $^3r$ represent the reward function for their respective subtasks. $^sr$ is then set as $^1r$, $^2r$, and $^3r$ based on the current stage.

			%
				%
	
	In reward coefficient matrix $R$, $d_{rb}$ represents the distance between the racket and the ball, while $d_{bt}$ represents the distance between the ball and the target. In addition, $v_{rhb_x}$ represents the velocity of the racket when it hits the ball, and $e_{lt}$ represents the error between the actual landing point of the ball and the target point.
	The reward function described above is primarily designed to achieve the core functionality of the robot.
	
	Additionally, there is a set of reward functions aimed at further improving the robot's performance, such as reducing jitter, minimizing energy consumption, and preventing collisions.
	These performance-enhancing reward functions are defined as follows.
	\begin{equation}
		^er(t)=c\sum_{i=0}^{7}T_i+d\sum_{i=0}^{7}(a_i(t)-a_i(t-1))^2 + e\sum_{i=0}^{N}F_c(i)
	\end{equation}
	In this equation, $^er$ represents the reward function for performance improvement, and $c$, $d$, and $e$ are its hyperparameters.
	$i$ denotes the joint index, and $T(i)$ represents the torque applied to each joint at the current time step.
	$a_i(t)$ and $a_i(t-1)$ refer to the joint actions given by the agent at the current and previous time steps.
	The final reward function can be expressed as follows:
	\begin{equation}
		r(t) = ^sr(t) +^er(t)
	\end{equation}
	\begin{algorithm}
		\caption{Curriculum reinforcement learning for the table tennis robot}
		\label{Reinforcement Algorithm}
		\begin{algorithmic}[1]
			\State Initialize parallel simulation environments, trajectory state buffer $\tau$, observation buffer $s_t$ and initial ball state buffer $B_{\tau}$
			\State Initialize the weights of Actor $\mu_\theta\left(s_t\right)$ \& Critic  $V_\phi\left(s_t\right)$ network with $\theta_0$ \& $\phi_0$ and training dataset $D$\
			\State Set training epochs $E_{num}=\left[E_1,\ E_2,\ E_3\right]$, horizon length $H$ 
			\For {$e$ in $E_{num}$ }
			\State Reset the count for parallel algorithm in observation
			\For {$epoch=1$, $e$}
			\For {$h$=1,$H$}
			\State Reproduce valid rakky trajectory with data from $B_\tau$ 
			\State Estimate table tennis ball state and obtain its trajectory state $\tau$ 
			\State Sample the output of actor $a_t\in\left[-1,1\right]$ with Gaussian Distribution $ N\left(\mu_\theta\left(Norm\left(s_t\right)\right),\sigma\right)$ and mapping $a_t$ to the joints' absolute position
			\State Control the robot to the absolute position by the PD controller and observe reward $^ir$ according to the reward function $^sr$, current stage $i$ and trajectory state $\tau$
			\State Calculate GAE $A_t$ and return $R_t$ and store $s_t$, $a_t$, $A_t$, $R_t$ in the training dataset $D$
			\EndFor
			\For {$m$=1,$M$}
			\State Sample minibatch data $s_t$,\ $a_t$,\ $A_t$,\ $R_t$ with batch size N from dataset D
			\State Update the weight of Actor $\mu_\theta\left(s_t\right)$ and Critic $V_\phi\left(s_t\right)$ by maximizing the PPO objective and regression on mean-squared error
			\EndFor
			\EndFor
			\EndFor
			
		\end{algorithmic}
	\end{algorithm}

	\subsection{Reinforcement Learning Algorithm \& Network Architecture}
	The PPO algorithm is selected for RL in our table tennis robot, and the actions are sampled from independent Gaussian distributions. During the training process, the robot should perform a variety of subtasks, therefore the learning rate cannot be applied to tricks such as warm-up, because it may prevent the robot from training. As an alternative, the learning rate is set to a constant value, and only when the training task enters the final stage does the learning rate gradually decrease. In this paper, the D2RL \cite{sinha2020d2rl} network is used as the network architecture, which concatenates the hidden layer with the observation and then feeds the concatenated tensor into the next layer. This network structure is mainly utilized to solve the problem that the layers in RL cannot be too deep. In addition, in our policy model, except for the last layer, the other several layers share the same network (including the observation). The network uses the parallel algorithm\cite{chan1982updating} for estimating the mean and variance of the observation to normalize the observation. Since the algorithm will gradually reduce the update amplitude of the estimated value, the counting number of the parallel algorithm should be reset to ensure that the agent can continue to learn in different stages. 
	
	When transitioning between training stages, the agent in the subsequent stage directly inherits the model weights learned from the previous stage. As a result, the same set of observations and actions is shared across different training stages. Given that the training tasks are defined in an order of increasing difficulty, this approach aligns closely with the principles of curriculum learning in continous learning. The primary objective of this method is to ensure that the agent retains the skills acquired in earlier stages, minimizing the risk of catastrophic forgetting.

	In the network, the observation consists of 37 dimensions, including the 7-dimensional joint position (7), joint velocity (7), racket position (7), racket posture (4), racket linear velocity (3), ball position (3), ball linear velocity (3), target location (2), trajectory state (1), and continuous trajectory state one-hot vector (4). The actor outputs a 7-dimensional tensor in the range of [-1,1]. This tensor can be mapped to the expected absolute joint position of the robot, which will be driven by the PD controller in the simulation environment. The output of the critic is the value (1), which is used to evaluate the advantage function that may be achieved by the actor. The method used in this article to train the hitting policy is summarized in Algorithm \ref{Reinforcement Algorithm}
	
	\subsection{Novelty}
	Compared to generating rally trajectories using offline data for training table tennis robots, the valid rally trajectory generation approach proposed in this paper is better suited for randomized environments. Introducing the impact of random Magnus force and air drag during the training process may lead to offline trajectories becoming invalid. Additionally, training with offline trajectories may result in overfitting issues, especially when the dataset is insufficient. Besides, the stage-based reward $^sr$ is proposed to decompose the final task and design different reward functions for different stages, allowing the agent to achieve the ability by completing several subtasks. The use of reward coefficient matrix $R$ to design the reward function in this paper offers several advantages. It allows for rewarding based on trajectory state changes during the hitting process and accelerates the computation of the reward function. This approach is not only applicable to table tennis robots but can also be employed in other reinforcement learning tasks. By leveraging matrices, the reward function becomes more flexible and adaptable, making it a versatile method applicable to a wide range of scenarios beyond just table tennis robots.
	
	\section{REAL2SIM TRANSFER}
	There are often significant discrepancies between simulation and reality, such as system delays and noise in data. To address these issues, different methods need to be employed to mitigate the effects caused by the disparities. In this paper, the focus is primarily on addressing the perception and execution systems of the robot. To bridge the gap between simulation and real-world scenarios,	various approaches have been taken, including data post-processing methods to accurately estimate the acquired data and introducing noise to the robot during the training process to enhance the model's generalization capabilities. 
	\subsection{Perception System}
	\begin{figure}[tbph]
		\centering
		\includegraphics[scale=0.6]{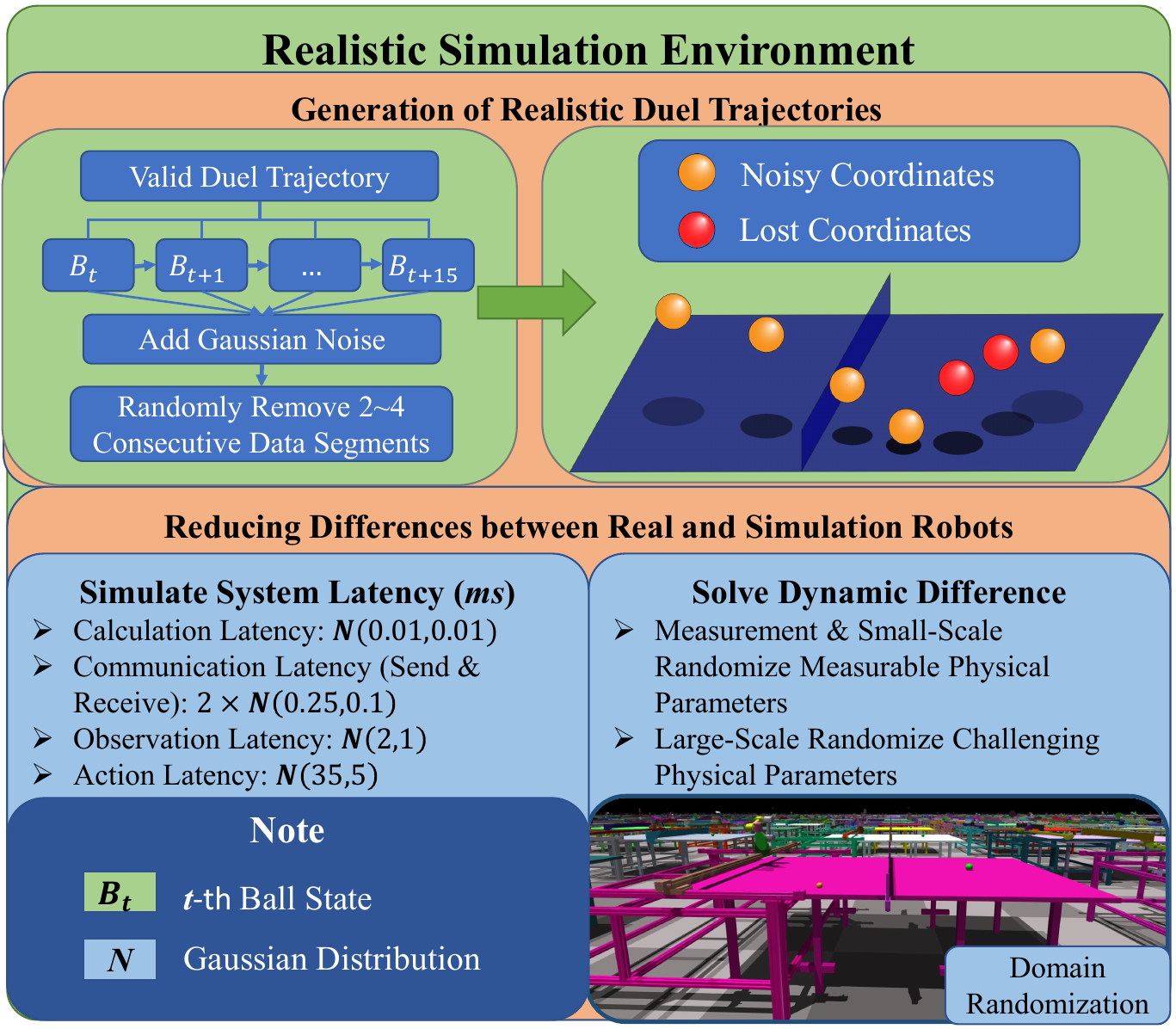}
		\caption{		To replicate the noise in binocular camera recognition of a table tennis ball, trajectory loss, and differences between real and simulated trajectories, we estimate delays in the table tennis robot system and simulates these disparities.}
		\label{fig:Simulation_environment.pdf}
	\end{figure}
	In the simulation environment, the position of the table tennis ball can be provided directly. Compared with the simulation, the real environment needs to track the table tennis ball through a camera and several algorithms to estimate the status of the table tennis ball. This method leads to increased delays and noise interference. In addition, the ball may disappear due to the image occlusion of the robot body or the limitation of the field of view (FOV) of the camera.  
	Considering that the robot does not encounter external noise interference in the simulated environment, Gaussian noise was added to the data  during the training process. 
	This approach helps the robot become more robust to potential real-world noise and uncertainties, allowing it to better generalize its learned behaviors to real-world scenarios. Similarly, random delays are also introduced into the robot's training process to simulate policy computation delays and communication delays. By incorporating these random delays and noises during training, the robot becomes more resilient and capable of handling real-world scenarios where such delays may occur. Details of the specific implementation method are illustrated in Figure \ref{fig:Simulation_environment.pdf}.
	\subsection{Decision and Execution System}
	\begin{figure}
		\centering
		\includegraphics[scale=0.7]{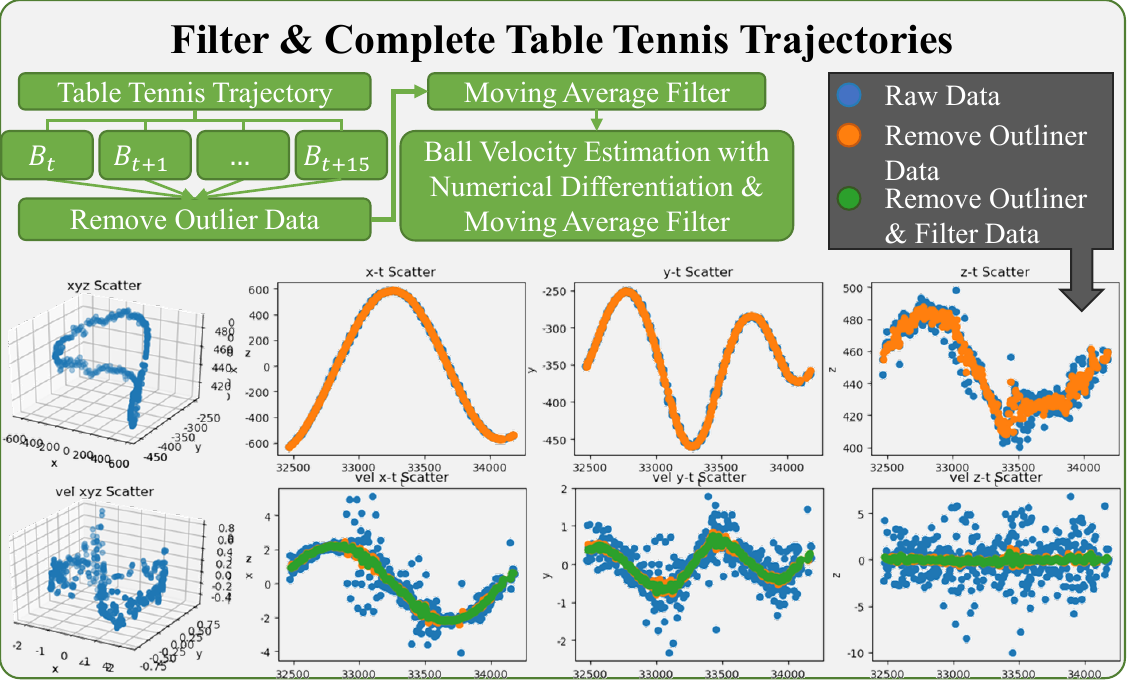}
		\caption{The left three-dimensional scatter plot shows the direct position and velocity data of the table tennis ball in 3D space obtained from a binocular camera. The right two-dimensional scatter plot displays position and velocity changes over time along the XYZ axes. Blue dots represent data directly from the camera, while other colors indicate data obtained after using a state estimation algorithm. The depicted trajectory corresponds to instances of a person swinging the ball, highlighting the algorithm's performance in capturing subtle position changes. The units for position, velocity, and acceleration are in millimeters (mm), meters per second (m/s), and meters per second squared (m/s²), respectively.}
		\label{fig:Filter Estimation}
	\end{figure}
	In addition to relying solely on reinforcement learning (RL) to solve all problems, state estimation algorithms are also applied in the Real2Sim process to enhance the credibility of the data acquired by the model. In the reality, our robots utilize a binocular camera to acquire the coordinates of the table tennis ball. However, this approach may introduce significant noise, delays, and even data loss in the obtained coordinates. To enhance the accuracy of the table tennis ball's state data, numerical differentiation and sliding window filtering are employed to estimate the current velocity of the table tennis ball and estimate the lost coordinates according to the previous table tennis ball's state. The table tennis trajectory obtained by using the above methods for the state estimation and trajectory completion is shown in Fig. \ref{fig:Filter Estimation}. These state estimation techniques help improve the accuracy and reliability of the data obtained from real-world interactions and measurements. By integrating state estimation with RL, the robot can obtain more reliable information, leading to more effective learning and better generalization in real-world scenarios. This combination of RL and state estimation plays a crucial role in achieving successful Real2Sim transfers and enhancing the overall performance of the robot.
	%
	\section{EXPERIMENTS}
	To verify our methods, we conducted comparative experiments to verify their effectiveness. In the experiments, the robot is required to hit the ball to a random target location on the opponent's court, and the trajectories of the incoming ball originate from random positions with random speeds (generated by the abovementioned valid rally trajectory generation).  In addition to training the robot to hit the ball within the simulated environment, the robot is also required to perform Real2Sim experiments. This entails the robot hitting table tennis ball trajectories introduced from real-world scenarios within the simulated environment. In essence, the robot must apply its learned skills from simulation to handle real-world scenarios involving ball trajectories with a similar level of proficiency. To complete the training of this policy, 40960 parallel environments are set up to generate training data and 122880 parallel environments are used to generate valid rally trajectories. 2 $\times$ A6000 graphics card are employed for the environment simulation and model training. The training time was approximately 5 hours.
	
	\subsection{Training Results in the Simulation Environment}
	
	\begin{table}[]
		\centering
		\caption{hyperparameters in reward functions}
		\label{hyperparameters table}
		\begin{tabular}{c|c|c|c|c|c|c|c|c}
		\hline\hline
		$a_{21}$ & $a_{22}$ & $a_{23}$ & $a_{31}$ & $a_{32}$&$a_{33}$ & $a_{41}$ & $a_{42}$ & $a_{43}$ \\ \hline
		1        & 0.25     & 0.1      & 10       & 4       & 1       & 1        & 0.25     & 0.1      \\ \hline\hline
		$a_{51}$ & $a_{52}$ & $a_{53}$ & $a_{63}$ & $a_{73}$ & $b_{73}$ &$c$  &$d$  &$e$  \\ \hline
		25       & 50       & 10       & 1        & 30       & 40       &0.02 &0.02 &0.1  \\ \hline
		\end{tabular}
	\end{table}

	To verify whether the stage-based reward $^sr$ and our collision model works, we conducted an experiment with 3 different policy models in the simulation environment and compared their results.
	In these policy models, the hyperparameters, network structure, and simulation environment are the same. The only differences between them are the differences in reward functions and whether a collision model tailored for spinning balls is incorporated.
    During training, the hyperparameters of the reward function are presented in Table \ref{hyperparameters table}.

	In the experiments, each episode refers to one round of the opponent and the robot playing against each other, which may be terminated prematurely due to illegal events (robot fails to catch the ball, the ball not crossing the net, robot's body touches the ball, etc.). The policy that incorporates the collision model refers to both training and testing the robot with the tailored collision model. Conversely, the policy that excludes the collision model implies that the training involves original collision model, while the testing utilizes the collision model tailored for spinning table tennis balls. This distinction highlights the influence of the tailored collision model on the robot's ability to handle spinning balls and demonstrates the necessity of its incorporation for comprehensive skill development and effective performance in real-world scenarios involving spin. 
	
\begin{figure}
		\centering
		\vspace{-1.5cm}
		\includegraphics[scale=0.3]{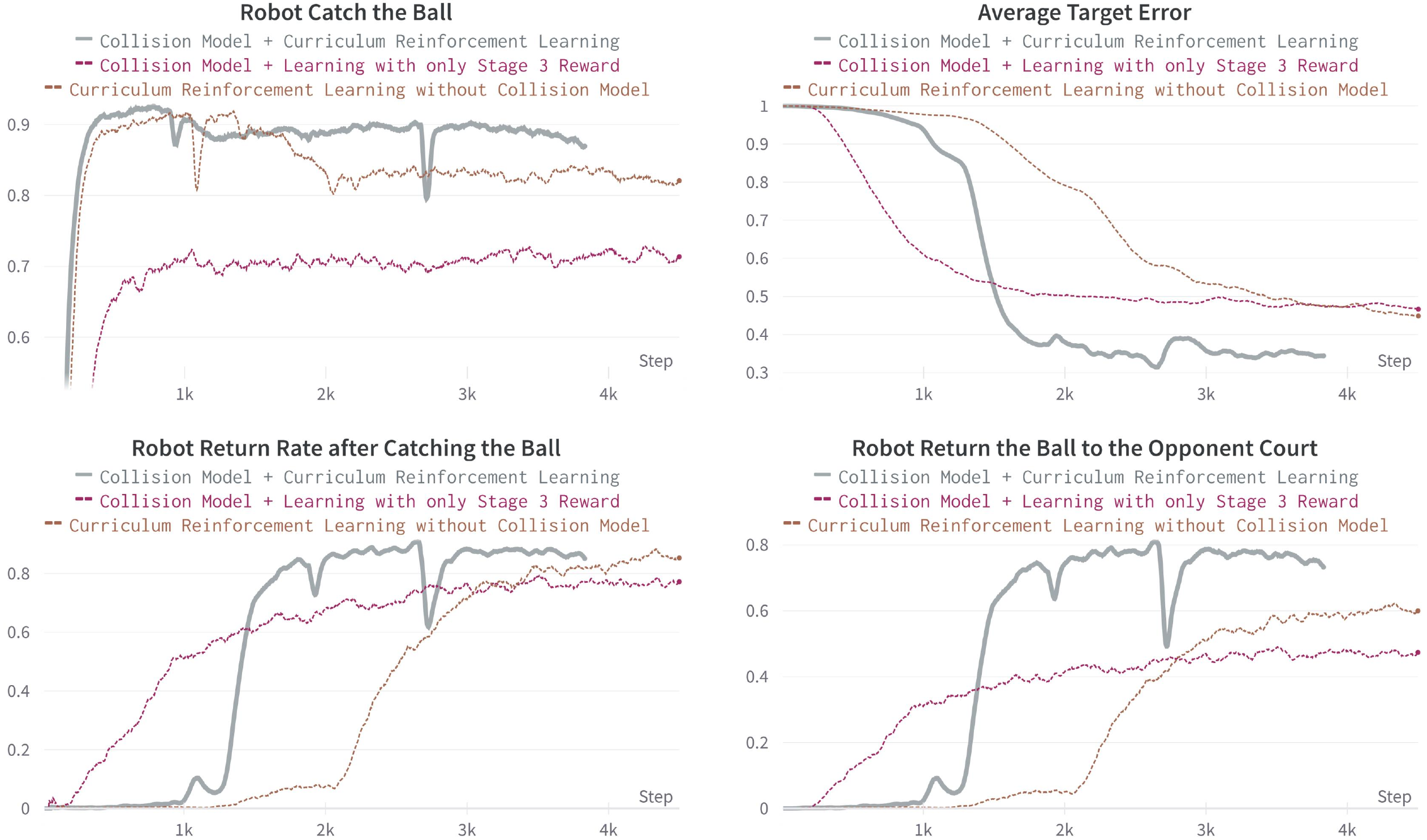}
		\caption{Training in Policy 1 (grey line) involves both curriculum reinforcement learning and the improved collision model. In contrast, Policy 2 (red line)  and Policy 3 (brown line)  exclusively incorporate the improved collision model and curriculum reinforcement learning  individually. In policy 1, each spike corresponds to the completion of a curriculum learning stage by the robot. However, the last peak which causes sharp decline in model performance, is not indicative of transitioning to the next stage. Instead, it stems from an unexplained model failure during the training process.}
		\label{fig:training result}
	\end{figure}

	The average target error, catching success rate, returning success rate, and return-to-opponent rate are provided in Fig. \ref{fig:training result}.
	
	From the figure, it is apparent that curriculum reinforcement learning combined with the improved collision model outperforms the scenario where only curriculum reinforcement learning is employed. Furthermore, if the robot is trained with only the final stage's reward and the collision model, the robot's ability to catch the ball will decrease significantly. The table tennis robot that underwent curriculum learning exhibits a proficient ability to catch spinning balls, regardless of whether spin was introduced during training. The only distinction is that the table tennis robot trained with both curriculum learning and spinning ball scenarios demonstrates a heightened capability to hit the ball effectively onto the opponent's table. The underlying cause for this phenomenon could be attributed to the fact that the table tennis robot subjected to spinning ball training is better to adapt to the trajectory changes resulting from the collision between the spinning ball and the robot's racket. 
	\subsection{Experiments with Table Tennis Trajectory in the Real2Sim}
\begin{figure}\centering
		\vspace{-1.5cm}
		\includegraphics[scale=0.7]{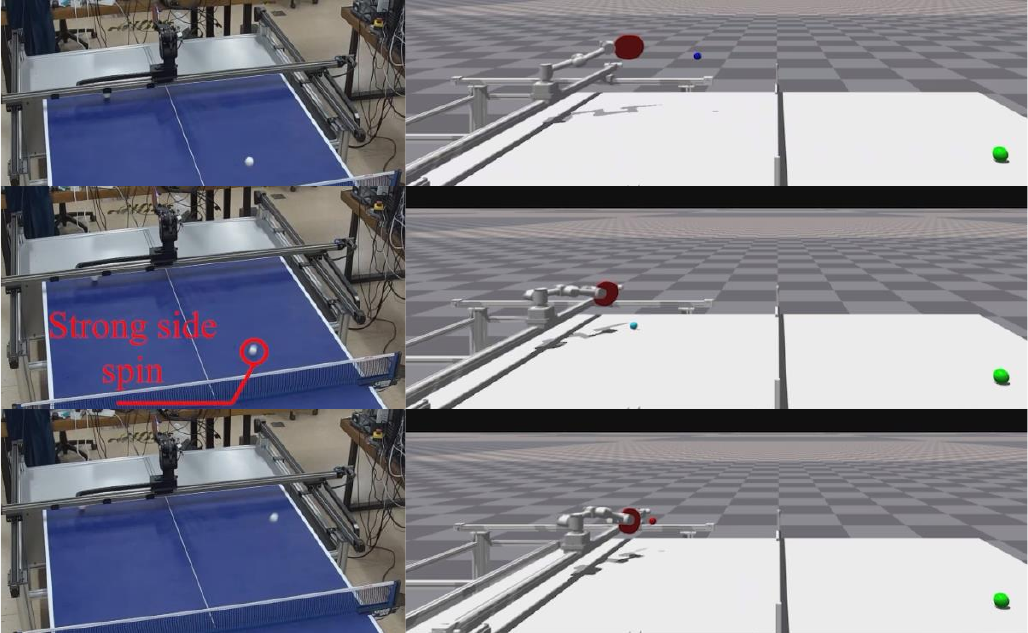}
		\caption{
			In the Real2Sim, the table tennis robot in the simulation environment is tasked with responding to the trajectory of the real-world ping-pong ball.
			The green ball represents the robot's target hitting point.
			The blue ball corresponds to $\tau_{0\rightarrow1}$ (the trajectory before the ball hits the robot's side of the table).
			The cyan ball represents $\tau_{1}$ (the moment the ball contacts the robot's side of the table).
			The red ball indicates $\tau_{3\rightarrow1}$ (the trajectory after the ball collides with the robot's paddle).
			In this experiment, our robot demonstrated a strong ability to handle spinning balls. However, due to space limitations, $\tau_{1\rightarrow2}$, $\tau_{2}$, and $\tau_{2\rightarrow3}$ are not depicted, which might mean the robot's full capability is not completely showcased.
		For a more detailed demonstration of the robot's behavior in the Real2Sim, please refer to the video linked in the abstract.}
		\label{fig:Real2Sim_Test}
	\end{figure}

	To evaluate the real-world performance of our robot, this study imports real-world table tennis ball trajectories into the simulation environment. 
	In the Real2Sim tests, the robot's performance is illustrated in Figure \ref{fig:Real2Sim_Test}.
	Through the Real2Sim transfer process, the robot's ability to handle spinning balls is assessed.
	The performance of different algorithm is shown in TABLE \Ref{table experiment}.  
	
	In TABLE \Ref{table experiment}, all table tennis trajectories (with the exception of Muscular Robot and Defensive Player) are generated by the oscillating ball launcher in reality or by random initial position and random initial velocity in simulation. A random target refers to whether the robot can hit the ball to a specified target location. The end-to-end model refers to a policy model that is independent of trajectory prediction or robotics algorithms.
	\begin{table}[]
		\centering
		\caption{the performance of different policy model}
		\label{table experiment}
		\begin{tabular}{c|c|c|c|c}
			\hline
			\begin{tabular}[c]{@{}c@{}}Algorithms \\ Name\end{tabular} & \begin{tabular}[c]{@{}c@{}}End-to\\-end \end{tabular} & \begin{tabular}[c]{@{}c@{}}Return\\ Rate\end{tabular}            & \begin{tabular}[c]{@{}c@{}}Return\\ after Catch\end{tabular} & \begin{tabular}[c]{@{}c@{}}Target \\ Error\end{tabular} \\ \hline\hline
			\begin{tabular}[c]{@{}c@{}}MLP+\\ ARS\cite{abeyruwan2023sim2real}\end{tabular}         & $\times$                                       & 88.60\%                                                          & -                                                    & 0.4m                                                    \\ \hline
			\begin{tabular}[c]{@{}c@{}}Muscular \\ Robot\cite{buchler2022learning}\end{tabular}  & \checkmark                                   & 75\%                                                            & 78.1\%                                                            & -                                         \\ \hline
			\begin{tabular}[c]{@{}c@{}}Defensive \\ Player\cite{liu2013racket}\end{tabular} & $\times$                                       & 80\%                                                             & -                                                    & -                                         \\ \hline
			\begin{tabular}[c]{@{}c@{}}Focused \\ Player\cite{liu2013racket}\end{tabular}   & $\times$                                       & 50\%                                                             & -                                                    & 0.25m                                                   \\ \hline
			\begin{tabular}[c]{@{}c@{}}8dof+GCBC\\ +SSP\cite{mulling2013learning}\end{tabular}   & \checkmark                                   & \begin{tabular}[c]{@{}c@{}}21\%\\ (\textless{}0.3m)\end{tabular} & -                                                    & 0.84m                                                   \\ \hline
			\begin{tabular}[c]{@{}c@{}}MLP+\\ PPO\cite{mahjourian2018hierarchical}\end{tabular}         & $\times$                                       & 40\%                                                             & -                                                    & -                                         \\ \hline
			\begin{tabular}[c]{@{}c@{}}CNN+\\ ES+ AF\cite{gao2020robotic}\end{tabular}       & \checkmark                                   & 79\%                                                             & 80.6\%                                                            & -                                         \\ \hline
			\begin{tabular}[c]{@{}c@{}} Collision Model \\\& Only Stage 3 \\ Training [our]\end{tabular}    & \checkmark                                   & 51\%                                                             & 84\%                                                                 & 0.44m                                                       \\ \hline
			\begin{tabular}[c]{@{}c@{}} Collision Model \\\& Stage 1+2+3 \\ Training [our]\end{tabular}    & \checkmark                                   & 81\%                                                             & 90\%                                         & 0.31m                                \\ \hline
		\end{tabular}
	\end{table}

	The data in the table clearly indicates that the agent trained using curriculum reinforcement learning demonstrates superior performance compared to the agent trained directly on the target task. Additionally, it can be observed that our algorithm exhibits a slight decrease in the success rate compared to the algorithm with the highest return rate. Moreover, it has the best success rate in returning after hitting the ball. In addition to the abilities mentioned above, our table tennis robot demonstrates the ability to hit the ball to arbitrary target points. This capability is reflected in our algorithm's nearly lowest target  error.
	
	\section{CONCLUSIONS}
	In this research paper, we have introduced a series of methodologies aimed at training a table tennis robot capable of playing against spinning balls. Through a combination of techniques, including curriculum reinforcement learning, improved collision modeling, trajectory state definition, effective trajectory collection, and Real2Sim transfer, we have demonstrated a comprehensive approach to enhance the robot's performance in handling challenging spinning ball scenarios. 
	
	With trajectory states defination and and the collection of valid rally trajectories, this study successfully accomplished data collection within a large-scale simulation environment. Through curriculum reinforcement learning, this paper successfully addressed the issue of the table tennis robot's ability to hit the ball to the target point. Furthermore, by introducing analysis of the table tennis ball's trajectory after collision, the paper enables the robot to handle spinning balls. To validate the robot's capability to handle spinning balls in real-world, the paper constructed a Real2Sim transfer scheme, importing strong spinning ball trajectories from the real-world into the simulation environment. The results of these experiments have been presented in the accompanying video.
	
	\bibliographystyle{elsarticle-num} 
	
	\bibliography{ref}
	
	
		
		
		
\end{document}